\documentclass{article}
\usepackage{subcaption}

\usepackage{microtype}
\usepackage{graphicx}
\usepackage{booktabs} 
\usepackage{enumitem}
\usepackage{tcolorbox}
\usepackage{hyperref}

\usepackage[accepted]{icml2025}

\usepackage{amsmath}
\usepackage{amssymb}
\usepackage{mathtools}
\usepackage{amsthm}

\usepackage[capitalize,noabbrev]{cleveref}

\theoremstyle{plain}
\newtheorem{theorem}{Theorem}[section]

\theoremstyle{definition}

\theoremstyle{remark}
\newtheorem{remark}[theorem]{Remark}

\usepackage[textsize=tiny]{todonotes}

\icmltitlerunning{Unveiling Hidden Pivotal Players with GoalNet: A GNN-Based Soccer Player Evaluation System}

\begin{document}

\twocolumn[
\icmltitle{Unveiling Hidden Pivotal Players with GoalNet: \\ A GNN-Based Soccer Player Evaluation System}



\icmlsetsymbol{equal}{*}

\begin{icmlauthorlist}
\icmlauthor{Jacky Hao Jiang}{yyy}
\icmlauthor{Jerry Cai}{yyy}
\icmlauthor{Anastasios Kyrillidis}{yyy,zzz}
\end{icmlauthorlist}

\icmlaffiliation{yyy}{Department of Computer Science, Rice University, Houston, USA}
\icmlaffiliation{zzz}{Ken Kennedy Institute, Rice University, Houston, USA}

\icmlcorrespondingauthor{Jacky Hao Jiang}{hj37@rice.edu}
\icmlcorrespondingauthor{Jerry Cai}{yc139@rice.edu}
\icmlcorrespondingauthor{Anastasios Kyrillidis}{anastasios@rice.edu}
\icmlkeywords{Machine Learning, ICML}

\vskip 0.3in
]



\printAffiliationsAndNotice{}  

\begin{abstract}
Soccer analysis tools emphasize metrics such as expected goals, leading to an overrepresentation of attacking players' contributions and overlooking players who facilitate ball control and link attacks. 
Examples include Rodri from Manchester City and Palhinha who just transferred to Bayern Munich.
To address this bias, we aim to identify players with pivotal roles in a soccer team, incorporating both spatial and temporal features.

In this work, we introduce a GNN-based framework that assigns individual credit for changes in expected threat (xT) \cite{singh2019introducing}, thus capturing overlooked yet vital contributions in soccer. Our pipeline encodes both spatial and temporal features in event-centric graphs, enabling fair attribution of non-scoring actions such as defensive or transitional plays. We incorporate centrality measures into the learned player embeddings, ensuring that ball-retaining defenders and defensive midfielders receive due recognition for their overall impact. Furthermore, we explore diverse GNN variants—including Graph Attention Networks and Transformer-based models—to handle long-range dependencies and evolving match contexts, discussing their relative performance and computational complexity. Experiments on real match data confirm the robustness of our approach in highlighting pivotal roles that traditional attacking metrics typically miss, underscoring the model’s utility for more comprehensive soccer analytics.

\end{abstract}

\section{Introduction}

\textbf{Background.}
Football player performance analysis has traditionally been conducted using prevalent metrics, such as goals, assists, chances created, and expected goals (xG). 
These are all good metrics for understanding the attacking contributions of players. 
Still, they often emphasize the contributions of forwards and playmakers, while sometimes underestimating the extremely central roles of players who typically work more profoundly in the field. 
At this point, players like defensive midfielders and central defenders bear the brunt of the responsibility in ball retention, transitions, and link-up play, that is a sine qua non of how a team performs. 
Despite their importance, their contributions are rarely reflected in conventional performance evaluations; thus, a gap in our understanding of their impact is created.

Take, for instance, Rodri, the holding midfielder at Manchester City, who facilitates the retention of possession very well and initiates quick transitions; he plays a pivotal role in linking defense with attack. 
Other midfielders include João Palhinha, who recently joined Bayern Munich from Fulham in a breakthrough deal and shines in the art of breaking opponent's attacks and starting forward play. 
These players thus contribute a lot to the team dynamics. 
Still, their contributions are hardly appreciated if we just focus on the number of goal-scoring or assisting actions. 

\textbf{Motivation.}
Recent advances give promising results in holistic evaluations of player performance. 
For instance, models like Value of Actions by Estimating Possession (VAEP) have demonstrated how machine learning can assign values to every player action by predicting their impact on scoring opportunities \cite{decroos2019vaep}. 
Similarly, Playerank by \cite{pappalardo2021playerank} used graph-based approaches to evaluate players based on their interactions and broader tactical context, extending evaluations beyond individual metrics. 
These methods illustrate the potential of leveraging advanced frameworks to uncover indirect contributions often overlooked by traditional metrics.

Graph neural networks (GNNs) \citep{kipf2017semi,velickovic2018gat} are seen as a particularly powerful framework for analyzing soccer matches since they are systems of player interactions \cite{wang2024tacticai}. 
If the model characterizes each player as a node in a system and interactions, such as passes, tackles, and movements, among other events, as edges, GNNs can capture and model the complexity in a game. 
Furthermore, since GNNs incorporate spatial and temporal features, they can make a more fine-grained sense of how individual players participate in the flow and eventual outcome of the game.

\textbf{Contributions.}
Building on the growing need to evaluate non-scoring roles and intricate interactions in soccer, we propose a GNN-based framework that assigns individual credit for changes in expected threat (xT) \cite{singh2019introducing}. 
Specifically, our contributions are as follows:\vspace{-0.25cm}
\begin{enumerate}[leftmargin=*]
    \item We develop a novel pipeline that constructs event-centric graphs encoding both spatial and temporal features, thereby capturing the complexity of soccer’s on-field dynamics and enabling the fair attribution of indirect contributions (e.g., defensive or transitional plays). \vspace{-0.2cm}
    \item We introduce a mechanism to combine GNN outputs with centrality measures, ensuring that facilitators such as defensive midfielders receive recognition proportional to their overall impact on the team’s performance. \vspace{-0.2cm}
    \item We present multiple GNN variants (including Graph Attention Networks and Transformer-based approaches) to illustrate how different architectures can model long-range player interactions and temporal sequences, and we offer an analysis of their relative merits and computational trade-offs. \vspace{-0.2cm}
    \item We validate our methodology on real match data, demonstrating its effectiveness in highlighting pivotal contributors beyond traditional attacking metrics. This evaluation shows the robustness of our approach in accurately quantifying roles such as ball retention, defensive organization, and transition initiation.
\end{enumerate}

\vspace{-0.1cm}
\section{Related work}
\vspace{-0.1cm}


\textbf{GNNs for player and team evaluation.}
In \cite{decroos2019vaep}, the authors introduced the concept of Value of Actions by Estimating Possession (VAEP), a framework that assigns values to every player action based on its likelihood of improving or diminishing a team’s chances of scoring or conceding. 
This approach allowed for more comprehensive evaluations of both attacking and defensive actions, particularly those that go unnoticed by traditional statistics such as goals and assists.

The model presented in \textit{Making Offensive Play Predictable} employs Graph Neural Networks (GNNs) to analyze defensive actions and their influence on offensive play. By modeling players as nodes and interactions (e.g., defensive pressure, passes) as edges, the model predicts pass recipients, shot probabilities, and defensive impact on attacking strategies \cite{stockl2021offensive}.

\cite{velickovic2021gat} demonstrated that using attention mechanisms within GNNs enables the model to assign varying importance to different player interactions, reflecting their influence on the game. 
This is useful in identifying playmakers who may not score or assist directly but are critical in creating opportunities \cite{velickovic2021gat}.

Moreover, Anzer et al. \cite{anzer2020tactics} demonstrated the potential of GNNs in detecting tactical patterns in soccer. 
Their semi-supervised model successfully identified frequent subgraph patterns representing tactical plays such as pressing formations or counterattacks. 
The ability to model team-level interactions and detect tactical setups gives GNNs a unique edge over traditional sports analytics models \cite{anzer2020tactics}.

\textbf{Valuing player's actions beyond goals.}
A key development in soccer analytics has been to extend the valuation of player actions beyond goals and assists. 
In work by Decroos et al. \cite{Decroos2019actions}, the Actions best capture this Speak Louder than Goals model, which assigns value to every player action by predicting the expected increase or decrease in the probability of scoring or conceding as a result of that action. 
This methodology allows for a better understanding of the contributions of players who are good at the build-up of play or defensively oriented \cite{Decroos2019actions}.
Similarly, the xT framework \citep{singh2019introducing} evaluates how actions change the probability of scoring over time. 
\textit{These methods inspire our approach, which integrates GNN-based embeddings with xT changes to highlight players' indirect yet vital roles.}

\textbf{Our novelty.} 
While previous work has analyzed players' influence using graph models or assigned values to actions, our approach unites these perspectives. 
We build upon the representational power of GNNs to distribute xT changes at the player level. 
Unlike methods focusing on final outcomes (e.g., goals) or static centrality measures, we incorporate the learned embeddings and temporal context to credit players who shape opportunities indirectly. 
Our contribution lies in generating more nuanced player valuations that capture complex, evolving interactions and highlight the hidden orchestrators of the game.



\section{GoalNet and Its Variants}


Every event is represented as a graph \( G = (V, E) \), where \( V \) denotes the set of players (nodes) and \( E \) represents the set of interactions (edges) between players. 
This graph-based representation serves as the foundation for all architectures discussed in this paper, including the basic GoalNet, Graph Attention Networks (GAT), and TransGoalNet (Graph Transformers). 

\subsection{Basic GoalNet (GNN)}
The Basic GoalNet architecture utilizes a basic GNN framework to analyze soccer match events. 

\textbf{Node attributes.}
Every node of the graph represents one player who participated in the current or past \( k \) events. The attribute set of a node \( v \in V \) is taken from the player-specific information in the dataset (see in Experiments).
Every node possesses the following $d$ attributes:

\begin{tcolorbox}[width=\linewidth,colback={green!10},title={Player's Statistics},colbacktitle=yellow!10,coltitle=black]   
\vspace{-0.15cm}
\begin{itemize}[leftmargin=*]
    \item \textit{Goals}: \# of goals scored by the player. \vspace{-0.2cm}
    \item \textit{Successful Dribbles}: \# of successful dribbles the player completes. \vspace{-0.2cm}
    \item \textit{Tackles}: \# of successful player tackles. \vspace{-0.2cm}
    \item \textit{Accurate Passes Percentage}: The percentage of successful passes. \vspace{-0.2cm}
    \item \textit{Rating}: The match rating assigned to the player. \vspace{-0.2cm}
    \item \textit{Goal Conversion Percentage}: The player’s efficiency in converting shots into goals. \vspace{-0.2cm}
    \item \textit{Interceptions}: \# of successful interceptions. \vspace{-0.2cm}
    \item \textit{Clearances}: \# of defensive clearances. \vspace{-0.15cm}
    \item \textit{Accurate Passes}: \# of accurate passes. \vspace{-0.15cm}
    \item \textit{Key Passes}: \# of  passes leading directly to a goal-scoring opportunity. \vspace{-0.15cm}
\end{itemize}
\end{tcolorbox}
Each node is thus characterized by a feature vector \( \mathbf{x}_v \) containing these player statistics.

\textbf{Edge attributes.}
Edges \( e \in E \) represent player interactions in the current event context, such as a pass, tackle, or other forms of player-to-player interaction. The edge attributes capture the nature of these interactions. The following $d'$ attributes are considered for each edge: \vspace{-0.2cm}
\begin{tcolorbox}[width=\linewidth,colback={green!10},title={Player's Interactions},colbacktitle=yellow!10,coltitle=black] 
\vspace{-0.15cm}
\begin{itemize}[leftmargin=*]
    \item \textit{Pass Information}: If the event involves a pass, we capture the pass's start and end locations, as well as the recipient of the pass. \vspace{-0.2cm}
    \item \textit{Event Type and Result}: These describe the type of event (e.g., pass, tackle) and its result (e.g., successful or unsuccessful). \vspace{-0.2cm}
    \item \textit{Spatial Coordinates}: The starting and ending coordinates of the ball, \( (x_{\text{start}}, y_{\text{start}}) \) and \( (x_{\text{end}}, y_{\text{end}}) \), which reflect the spatial dynamics of the event. \vspace{-0.2cm}
    \item \textit{xT Value and Change}: The expected threat (xT) value of the event, including the change in xT (\texttt{xT\_change}) as a result of the interaction. The xT metric quantifies how an action changes the probability of eventually scoring. 
    Following \cite{singh2019introducing}, xT is computed by dividing the pitch into zones and estimating the likelihood of scoring from each zone. 
    The difference \(\Delta xT\) represents how much an action increases or decreases this probability.
 \vspace{-0.2cm}
\end{itemize}
\end{tcolorbox}
Overall, each edge \( e_{uv} \) connecting player \( u \) and player \( v \) is described by a feature vector \( \mathbf{e}_{uv} \):
\[
\mathbf{e}_{uv} = \left[ \begin{split}
\text{Pass Information}, \text{Event Type and Result} \\
 (x_{\text{start}}, y_{\text{start}}),(x_{\text{end}}, y_{\text{end}}) \\ 
 \text{xT Value}, \text{xT Change}
\end{split} \right].
\]

\textbf{Graph representation and the GoalNet architecture.}
In each soccer's graph \( G = (V, E) \), the node feature matrix \( \mathbf{X} \in \mathbb{R}^{|V| \times d} \) and the edge feature matrix \( \mathbf{E} \in \mathbb{R}^{|E| \times d'} \) are the inputs to our graph neural network (GNN), where \( d \) and \( d' \) represent the dimensionality of the node and edge feature vectors, respectively.

\begin{figure*}[ht]
\centering
\includegraphics[width=\textwidth]{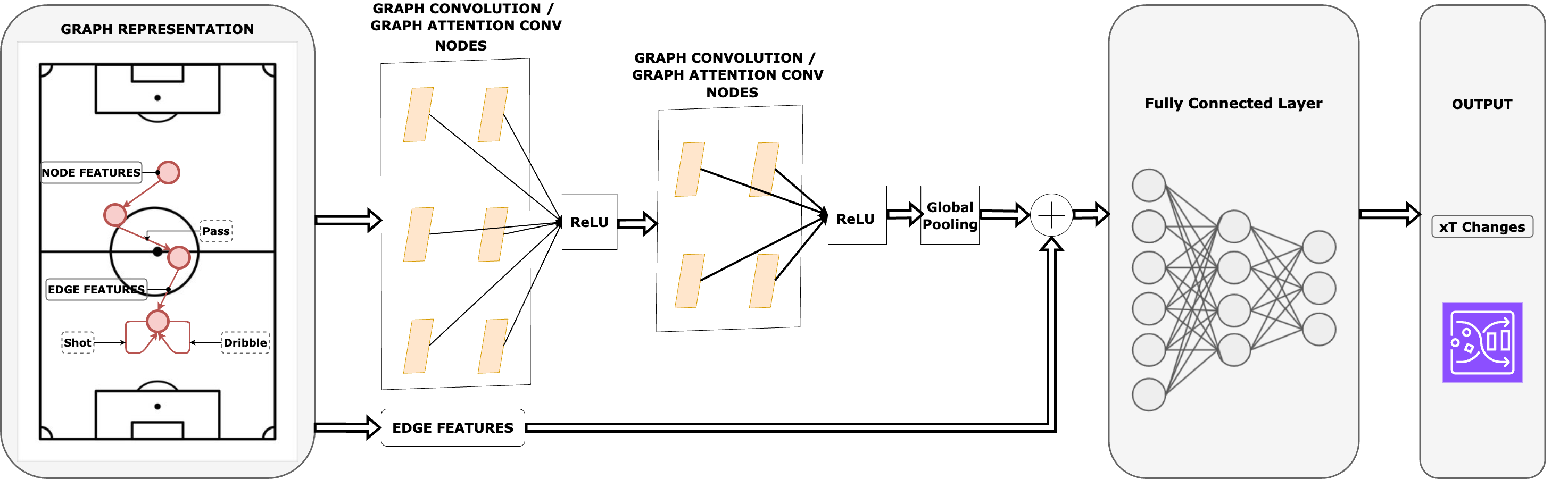}  
\caption{Model Visualization: Graph representing interactions and players will be fed to Graph convolution blocks and fully connected layer for prediction.}
\label{fig:histogram1}
\end{figure*}

The core of our model is based on a Graph Convolutional Network (GCN) that processes both node and edge features, enabling the model to learn from the structure of player interactions during match events. 
We employ GCN basic blocks in GoalNet to generate meaningful node embeddings that capture the influence and importance of players involved in each event. 
Classic centrality metrics from graph theory \cite{freeman1991centrality}, such as degree centrality, betweenness centrality, and closeness centrality \cite{zhang2017degree}, are helpful to measure a player's role in static networks. 
Instead, we make use of the learned node embeddings from the GNN to dynamically distribute the values of expected threat (xT) to the players based on their involvement in events taking place in the match.
Once the model predicts \(\Delta xT\), we distribute this value across involved players using learned node embeddings. Higher embedding magnitudes correspond to greater player contribution. By normalizing embedding magnitudes, we proportionally allocate \(\Delta xT\) among players.
Overall, the GNN model consists of the following components:

---\textit{Graph Convolution Layers.} 
Our model employs two layers of graph convolutions. 
Let \( \mathbf{H}^{(l)} \) represent the hidden node feature matrix at layer \( l \), and \( \mathbf{A} \) be the adjacency matrix of the graph. The forward pass at each layer is computed as:
\[
\mathbf{H}^{(l+1)} = \sigma(\mathbf{A} \mathbf{H}^{(l)} \mathbf{W}^{(l)}),
\]
where \( \sigma \) is the activation function (ReLU in our case) and \( \mathbf{W}^{(l)} \) is the trainable weight matrix for layer \( l \).

---\textit{Edge Feature Processing.}
In addition to node features, edge features are processed through a multi-layer perceptron (MLP) to learn representations for player interactions. The MLP consists of two fully connected layers with ReLU activations. For each edge, let \( \mathbf{e}_{uv} \) represent the edge feature vector between nodes \( u \) and \( v \). The MLP transforms \( \mathbf{e}_{uv} \) as:
\[
\mathbf{e}'_{uv} = \text{ReLU}(\mathbf{W}_2 \cdot \text{ReLU}(\mathbf{W}_1 \cdot \mathbf{e}_{uv})),
\]
where \( \mathbf{W}_1 \) and \( \mathbf{W}_2 \) are the weight matrices for the MLP layers. The processed edge features are then incorporated into the GNN to enrich the representation of player interactions.

---\textit{Global Mean Pooling.}
A global mean pooling layer is applied to aggregate node embeddings into a single graph-level embedding. This step ensures that the contributions of all players in the event are considered when predicting the xT change. The global pooling operation is defined as: $\mathbf{z} = \frac{1}{|V|} \sum_{v \in V} \mathbf{h}_v$,
where \( \mathbf{h}_v \) is the hidden representation of node \( v \) after the graph convolutions, and \( |V| \) is the number of nodes (players) in the graph.

---\textit{Fully Connected Layers.}
The pooled graph embedding \( \mathbf{z} \) is passed through two fully connected layers with ReLU activations to predict the final output, the xT change for the event. The output of the model is a scalar value \( \hat{y} \), representing the predicted xT change:
\[
\hat{y} = \mathbf{W}_4 \cdot \text{ReLU}(\mathbf{W}_3 \cdot \mathbf{z}),
\]
where \( \mathbf{W}_3 \) and \( \mathbf{W}_4 \) are the weights of the fully connected layers.

The above follow classical GNN implementations. 
The entire model is trained using the Mean Squared Error (MSE) loss function to minimize the difference between the predicted xT changes and the actual xT changes. 

\begin{remark}
{
\it The goal of training GoalNet should not be computing a more accurate xT. 
Rather, it is to attach this xT to individual players based on what they have learned from the representations (embeddings) in the graph. 
These embeddings encode information about how the player interacts, contributes to ball progression, and affects game dynamics: Players with higher embedding magnitudes receive a more significant share of the xT, as their learned representation suggests a stronger contribution to the event's outcome.
The GoalNet model learns these representations by performing multi-layer convolutional operations capturing spatial and temporal dependencies in the match events. 
When the model trains, node embeddings can be extracted to distribute the xT value for each event proportionally. 

Integrating node embeddings as a basis for xT distribution allows for a more accurate and fair assessment of player contributions. Players who are involved in critical transitions or build-up play but may not be directly associated with goal-scoring opportunities are appropriately credited for their impact on the match. This leads to a more holistic evaluation of player performance, which is particularly beneficial for defensive midfielders and central defenders, who play pivotal roles in facilitating team dynamics.
}
\end{remark}

\textbf{Temporal Context.}
The temporal aspect of soccer matches is critical in understanding player performance and interactions. 
To accurately model player contributions, it is important to capture the spatial relationships between players and the sequence of events leading up to each match situation. 
Our approach incorporates temporal context by integrating information from preceding events into the graph representation used by GoalNet.

Several prior works, such as VAEP \cite{decroos2019vaep} and Playerank \cite{pappalardo2021playerank}, have explored temporal dependencies to evaluate player actions. 
These approaches primarily treat events sequentially or as isolated instances, rather than leveraging graph-based frameworks with temporal weighting. 
Our method combines graph neural networks with a dynamic integration of spatial and temporal features, enabling a nuanced understanding of player contributions over time. 


\begin{remark}
{
\it For each event, we include the previous $k$ events in the graph construction. This ensures the graph captures recent interactions and player involvement, allowing the model to understand how prior plays influence the current game state. The choice of \( k \) is determined based on empirical performance, balancing between capturing relevant temporal information and maintaining computational efficiency. 
We conduct ablation studies to evaluate the impact of varying \( k \) values on model performance, as detailed in Appendix A.
}
\end{remark}

\begin{remark}
{
\it Each edge in the graph is associated with not only the spatial features (such as pass frequency and distances) but also temporal features, including the time elapsed since the start of the match and the time difference between successive events. These features provide a richer representation of the temporal flow of the game, helping to capture how quickly actions unfold and how the timing of events affects player outcomes.
}
\end{remark}

\begin{remark}
{
\it By incorporating temporal context, we allow the model to learn some sort-of ``causality'' in player interactions. For instance, a defensive interception may trigger a counterattack, where a sequence of quick passes leads to a scoring opportunity. 
Our approach ensures that these are captured by explicitly encoding the chronological order of events in the graph structure and incorporating temporal differences as features. 
The temporal sequence is maintained during graph construction, ensuring that the relationships between defensive and offensive contributions are captured effectively. 

}
\end{remark}

\subsection{GATGoalNet (Graph Attention Networks)}
GATGoalNet extends the GoalNet architecture by replacing standard graph convolutional layers with Graph Attention Network (GAT) layers \cite{velickovic2018gat}. 
This modification introduces attention-based mechanisms that dynamically weight player interactions, enabling the model to focus on the most influential relations in the game’s passing network and event dynamics.

\textbf{Attention mechanism.}
For each node \( v \) and its neighbor \( u \in N(v) \), GAT computes an attention coefficient \( \alpha_{uv} \) that determines the influence of \( u \)’s features on \( v \)’s update. This coefficient is computed as:
\[
\alpha_{uv} = \tfrac{\exp\left(\text{LeakyReLU}\left(\mathbf{a}^\top [\mathbf{W} \mathbf{h}_u^{(l)} \| \mathbf{W} \mathbf{h}_v^{(l)}]\right)\right)}{\sum_{w \in N(v)} \exp\left(\text{LeakyReLU}\left(\mathbf{a}^\top [\mathbf{W} \mathbf{h}_w^{(l)} \| \mathbf{W} \mathbf{h}_v^{(l)}]\right)\right)},
\]
where \( \mathbf{h}_u^{(l)} \) is the embedding of node \( u \) at layer \( l \), \( \mathbf{W} \) is a learnable projection matrix, \( \mathbf{a} \) is the shared attention vector, and \( \| \) denotes concatenation.

\textbf{Multi-head attention.}
To improve robustness and capture diverse interaction patterns, GATGoalNet employs multi-head attention. Each attention head independently computes coefficients \( \alpha_{uv}^{(m)} \) and corresponding embeddings, which are concatenated to form the final node representation:
\[
\mathbf{h}_v^{(l+1)} = \sigma\left(\bigg\Vert_{m=1}^M \sum_{u \in N(v)} \alpha_{uv}^{(m)} \mathbf{W}^{(m)} \mathbf{h}_u^{(l)}\right),
\]
where \( M \) is the number of attention heads, and \( \sigma \) is the activation function (e.g., ReLU).

A comprehensive breakdown of the architecture and technical considerations is provided in Appendix B.

\subsection{TransGoalNet (Graph Transformers)}
TransGoalNet extends the GoalNet framework by incorporating Graph Transformers \cite{dwivedi2020generalization}, which enable global attention across all nodes in a graph. Unlike GATGoalNet, which focuses on neighbor-based attention, Graph Transformers capture long-range dependencies and global structural patterns, offering a holistic view of player interactions. This capability is especially relevant in soccer, where off-ball positioning and distant player roles can significantly influence expected threat (xT) dynamics.

TransGoalNet replaces traditional GCN or GAT layers with Graph Transformer layers, enabling global attention and interaction modeling beyond local neighborhoods. This captures long-range dependencies and player dynamics.

\textbf{Positional and structural encodings.} Graph Transformers require positional information to interpret graph structures. In TransGoalNet, positional encodings are derived from player roles (e.g., defender, midfielder) and spatial coordinates on the pitch. These encodings are added to node features to provide contextual information about player positioning and roles.

\textbf{Relational encodings for edge features.} To integrate edge features, TransGoalNet uses relational encodings that influence attention coefficients. For each node pair \((u, v)\), the edge features \( \mathbf{e}_{uv} \) are transformed into a relational embedding \( \mathbf{r}_{uv} \), which modifies the self-attention mechanism:
\[
\alpha_{uv} = \frac{\exp\left(\frac{\mathbf{q}_u \cdot \mathbf{k}_v + \mathbf{r}_{uv}}{\sqrt{d}}\right)}{\sum_{w \in V} \exp\left(\frac{\mathbf{q}_u \cdot \mathbf{k}_w + \mathbf{r}_{uw}}{\sqrt{d}}\right)},
\]
where \( \mathbf{q}_u, \mathbf{k}_v \) are the query and key vectors derived from node embeddings, and \( \mathbf{r}_{uv} \) represents the relational encoding for the edge \((u, v)\).

\textbf{Self-attention mechanism.} Each transformer layer computes query, key, and value matrices \( (\mathbf{Q}, \mathbf{K}, \mathbf{V}) \) from the node embeddings \( \mathbf{H}^{(l)} \) at layer \( l \):
\[
\mathbf{Q} = \mathbf{H}^{(l)} \mathbf{W}_\text{Q}, \quad \mathbf{K} = \mathbf{H}^{(l)} \mathbf{W}_\text{K}, \quad \mathbf{V} = \mathbf{H}^{(l)} \mathbf{W}_\text{V}.
\]
Attention is computed globally across all nodes, ensuring that distant players and off-ball movements are included in the representation.

\textbf{Layer normalization and feed-forward networks.} Each transformer layer includes layer normalization and a position-wise feed-forward network to stabilize training and introduce non-linear transformations:
\[
\mathbf{H}^{(l+1)} = \text{LayerNorm}\left(\mathbf{H}^{(l)} + \text{Att}(\mathbf{Q}, \mathbf{K}, \mathbf{V})\right),
\]
\[
\mathbf{H}^{(l+1)} = \text{LayerNorm}\left(\mathbf{H}^{(l+1)} + \text{FFN}(\mathbf{H}^{(l+1)})\right),
\]
where \(\text{FFN}\) is a feed-forward network applied independently to each node.


\textbf{xT Attribution to Players.}
To distribute \( \Delta xT \) among players, we compute the magnitude of each node embedding \( \|\mathbf{h}_v^{(L)}\| \) and normalize it across all nodes:
\[
\text{xT}_v = \frac{\|\mathbf{h}_v^{(L)}\|}{\sum_{u \in V} \|\mathbf{h}_u^{(L)}\|} \cdot \Delta xT.
\]
This ensures that both direct interactions and global strategic contributions are reflected in the attribution.



Appendix C provides an extensive analysis of the design choices and their implications in our framework.

\section{Experiments}
\textbf{Dataset.}
This study utilizes two primary datasets: event data and player-specific data.
The event data is sourced from the publicly available StatsBomb Open Data repository \cite{statsbomb2018open}, specifically from the Premier League 2015/2016 season. 
This dataset provides detailed event-level information such as body part used, event type, event outcome, and associated spatial coordinates. 
\begin{tcolorbox}[width=\linewidth,colback={green!10},title={Event Data},colbacktitle=yellow!10,coltitle=black]   
\vspace{-0.15cm}
\begin{itemize}[leftmargin=*]
    \item \textbf{Number of Matches:} 380 matches from the 2015/2016 Premier League season. \vspace{-0.2cm}
    \item \textbf{Number of Events:} 758426 events, encompassing passes, tackles, shots, and other actions. \vspace{-0.2cm}
    \item \textbf{Number of Players:} 547 players are represented across all teams in the league. \vspace{-0.2cm}
    \item \textbf{Graph Parameters:} For each event, the graph \( G = (V, E) \) has: \vspace{-0.2cm}
    \begin{itemize}[leftmargin=*]
        \item \textbf{Nodes (\( V \)):} Up to 22 nodes, representing the 22 players on the field at the time of the event.
        \item \textbf{Edges (\( E \)):} Varies per event, representing player interactions such as passes or tackles.
        \item \textbf{Node Features (\( d \)):} Each node is described by \( d = 10 \) features, including goals, successful dribbles, tackles, accurate pass percentage.
        \item \textbf{Edge Features (\( d' \)):} Each edge has \( d' = 5 \) features, such as spatial coordinates, event type, and xT value.
    \end{itemize}
\end{itemize}
\end{tcolorbox}

\textbf{Player-Specific Data.}  
Player-specific data is collected from Sofascore, a widely used platform for soccer analytics. This dataset includes aggregated statistics and metrics for individual players: $i)$ accurate pass percentage, $ii)$ key passes, $iii)$ goal conversion percentage, $iv)$ match ratings, and $iv)$ defensive metrics like interceptions and clearances.
This data complements the event data by providing a comprehensive view of player performance over the season.

\textbf{Dataset Size and Coverage.}  
The combination of StatsBomb and Sofascore datasets allows us to represent diverse scenarios, including: $i)$ Different types of matches (home vs. away), $ii)$ player performances across positions and roles, and $iii)$ varied tactical setups and strategies.

\textbf{Data preprocessing.}
We transform the event data into SPDAL \cite{Decroos2019actions} format, a structured format developed to represent soccer matches as a sequence of on-the-ball actions. 
Each action in a game (pass, shot, dribble) is represented by 12 attributes, including the player's ID, team, action type, start and end positions, possible receiver, and the result. 
Player-specific data is processed every 90 minutes and normalized to all features.

\textbf{Response variable definition.}
This study uses the change Expected Threat (xT) \cite{singh2019introducing} as the response variable since it is designed to evaluate better how players and teams contribute to creating goal-scoring opportunities. Specifically, we have two ways of computing the xT changes:
\[
\Delta xT =
\begin{cases}
xT_{\text{current}} - xT_{\text{previous}}, & \text{same team}, \\
xT_{\text{current}} + xT_{\text{previous}}, & \text{different teams}.
\end{cases}
\]
If two events are from the same team, we want to know how much more threat the current event adds; if they are from different teams, we need to account for the current event erasing the opponent's xT and contributing to their own.

\textbf{Training details.}
The GoalNet framework was evaluated across three architectures: Basic GNN, GATGoalNet, and TransGoalNet. Each architecture was trained using the Adam optimizer \cite{kingma2014adam} with an initial learning rate of \(1 \times 10^{-4}\), incorporating a weight decay of \(1 \times 10^{-4}\) to prevent overfitting. The training process was conducted over 25 epochs, with the dataset split into 80\% for training and 20\% for validation. A batch size of 64 was used across all architectures, balancing computational efficiency and memory constraints.

Weight initialization was performed using the Xavier method \cite{glorot2010understanding} for linear layers and Kaiming initialization \cite{he2015delving} for graph convolution and attention layers. 
A learning rate scheduler \cite{smith2017cyclical} reduced the learning rate by 0.5 every 10 epochs. 
Early stopping \cite{prechelt1998automatic} was employed to prevent overfitting, halting training if the validation loss did not improve for 5 consecutive epochs.
The choice of batch size, learning rate, and number of layers was fine-tuned based on preliminary experiments to ensure robust training and generalization across validation data. 

\textbf{Evaluation Metrics.}
The performance of the GoalNet model was evaluated using two regression metrics: Mean Squared Error (MSE) and Mean Absolute Error (MAE). 

\begin{table*}[!htp]
\centering
\begin{tabular}{|l|l|l|l|l|}
\hline
\textbf{\#} & \textbf{VAEP} & \textbf{Baseline} & \textbf{GAT} & \textbf{TransGoalNet} \\ \hline
1 & Ryan Bennett (NC) & Andrew Surman (AFCB) & Mesut Özil (Ar) & Mesut Özil (Ar) \\ \hline
2 & Andrew Surman (AFCB) & Francesc F\'abregas (Ch) & Francesc F\'abregas (Ch) & Francesc F\'abregas (Ch) \\ \hline
3 & Harry Arter (AFCB) & Mesut Özil (Ar) & Andrew Surman (AFCB) & Junior Stanislas (AFCB) \\ \hline
\end{tabular}
\caption{Top Overall Players Across Models. NC = Norwich City; AFCB = AFC Bournemouth; Ch = Chelsea; Ar = Arsenal.}
\label{tab:overall_top_players}
\end{table*}

\begin{table*}[!htbp]
\centering
\begin{tabular}{|l|l|l|l|l|}
\hline
\textbf{Team} & \textbf{VAEP} & \textbf{Baseline} & \textbf{GAT} & \textbf{TransGoalNet} \\ \hline
AFC Bournemouth & Andrew Surman & Andrew Surman & Andrew Surman & Junior Stanislas \\ \hline
Arsenal & Aaron Ramsey & Mesut Özil & Mesut Özil & Mesut Özil \\ \hline
Aston Villa & Rudy Gestede & Idrissa Gana Gueye & Idrissa Gana Gueye & Rudy Gestede \\ \hline
Chelsea & Thibaut Courtois & Francesc F\'abregas & Francesc F\'abregas & Eden Hazard \\ \hline
Crystal Palace & Glenn Murray & Yohan Cabaye & Yohan Cabaye & Glenn Murray \\ \hline
\end{tabular}
\caption{Top Players by Team Across Models.}
\label{tab:team_top_players}
\end{table*}

\section{Results}
Here, we demonstrate the predictive accuracy, efficiency, and robustness of each model in capturing player contributions and event dynamics. 
For ablation studieson the impact of temporal features ($k$-values), please see the Appendix.

\subsection{Overall Player Ranking and Team-wise Top Players: Model Comparison}


\textbf{Top Overall Players Across Models.}
Table~\ref{tab:overall_top_players} presents the top overall players across different models. The VAEP model primarily ranks forwards highly due to its emphasis on goal-scoring and assist-related contributions. Conversely, the Baseline, GAT, and TransGoalNet models highlight midfielders and defenders, capturing their contributions to ball progression, defensive actions, and transitions—areas often undervalued in traditional metrics.

\textbf{Top Players by Team.}
Table~\ref{tab:team_top_players} presents the top player for each team across all models. While VAEP tends to rank attacking players highly, the Baseline, GAT, and TransGoalNet models incorporate defensive and midfield contributions, capturing a more holistic view of player impact.

\textbf{Comparison and Analysis of Results.}
The VAEP model predominantly highlights forwards due to its reliance on metrics associated with goal-scoring and assists. In contrast, the Baseline, GAT, and TransGoalNet models emphasize: $i)$ xThreat (xT) contributions, rewarding players involved in build-up play, transitions, and defensive actions (\textbf{Baseline Model}); $ii)$ player interactions, emphasizing passing, spatial positioning, and defensive contributions (\textbf{GAT Model}); and, $iii)$ long-range dependencies and temporal dynamics, ensuring a holistic evaluation of both direct and indirect contributions (\textbf{TransGoalNet Model}).

Overall, our results underline the superiority of graph-based models (GAT and TransGoalNet) in uncovering the contributions of midfielders and defenders, who play critical roles in ball retention, transition play, and defensive stability. By integrating player interactions and temporal dynamics, these models provide a more balanced evaluation of player impact compared to traditional metrics.
For instance, Andrew Surman (AFC Bournemouth) is highlighted as a top player in the Baseline and GAT models for his defensive and midfield contributions, which are undervalued by VAEP. Also, Mesut Özil (Arsenal) receives higher ranks in the Baseline and GAT models due to his involvement in creating xThreat across multiple phases of play, not just in the final third.

\subsection{Case Study}
This is Match 26 of the Bundesliga season, Freiburg vs. Bayer Leverkusen, which ended in a 3–2 victory for Leverkusen. The diagram shows a series of events leading to Patrik Schick's astonishing finish: \vspace{-0.15cm}
\begin{enumerate}[leftmargin=*]
    \item Granit Xhaka’s Forward Pass: Xhaka spots space behind the first and second lines of Freiburg’s defense. By playing a forward pass into that space, he disrupts Freiburg’s shape and opens up room for the attackers to operate in a relatively unpressured environment. \vspace{-0.15cm}
    \item Adam Hložek’s Dribble/Pass: Freed by Xhaka’s initial pass, Hložek has time and space to combine with Jeremie Frimpong. \vspace{-0.15cm}
    \item Jeremie Frimpong’s Cross: Frimpong then delivers a cross into the middle, setting up Patrik Schick. \vspace{-0.15cm}
    \item Patrik Schick’s Shot: Schick finishes off the move, scoring the goal. \vspace{-0.15cm}
\end{enumerate}

\begin{figure*}[ht]
    \centering
    \begin{subfigure}{1\textwidth}
        \centering
        \includegraphics[width=\textwidth]{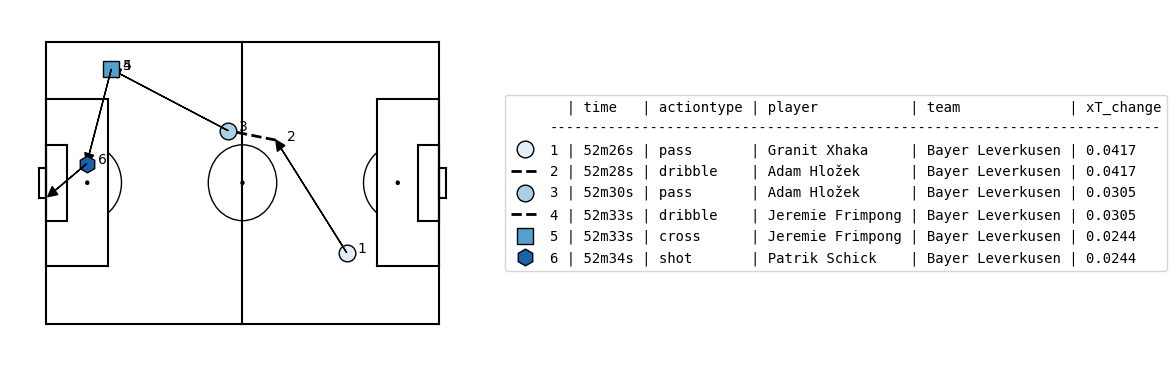}
        \caption{xT Changes of our model}
        \label{fig:model_pred}
    \end{subfigure}
    \hfill \vspace{-0.3cm}
    \begin{subfigure}{1\textwidth}
        \centering
        \includegraphics[width=\textwidth]{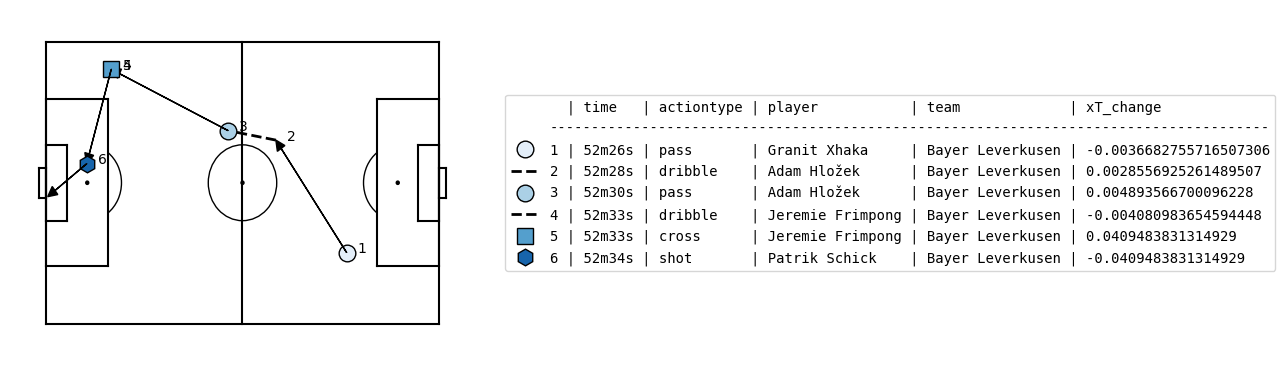}
        \caption{xT Changes for basic XT model}
        \label{fig:baseline_pred}
    \end{subfigure}
    \label{fig:two_column_fig} 
    \vspace{-0.5cm}
\end{figure*}
\textbf{Analysis.}
While the baseline and traditional model might focus on the final pass or cross (Hložek’s or Frimpong’s), the crucial action is the initial forward pass that breaks the defensive shape, which in this case is Xhaka's pass. 
In the second visualization, our model attributes higher expected threat (xT) change to Xhaka’s pass, reflecting that it set the stage for all subsequent actions.

\section{Conclusion}
The value of this work lies on the interesting case we consider: We introduce GoalNet, a GNN-based soccer player evaluation framework, designed to address the limitations of traditional metrics that predominantly favor attacking players. 
By incorporating spatial and temporal dynamics through graph-based representations, our approach effectively quantifies indirect but crucial contributions made by midfielders and defenders.

Our comparative analysis across different evaluation models---VAEP, Baseline, GAT, and TransGoalNet---revealed critical insights into player impact. Traditional approaches like VAEP emphasize goal-scoring and assist metrics, often leading to a bias in favor of attacking players. 
In contrast, our graph-based models, particularly GAT and TransGoalNet, successfully highlight the importance of midfielders and defenders by capturing their role in transitions, ball progression, and defensive stability. Key findings include: \vspace{-0.15cm}
\begin{itemize}[leftmargin=*]
    \item \textit{Midfielders and Defenders Recognition:} Unlike VAEP, which ranks mostly forwards at the top, GAT and TransGoalNet reward players contributing to build-up play, defensive interventions, and space creation. \vspace{-0.15cm}
    \item \textit{Superior Player Ranking Evaluation:} The results show that Andrew Surman (AFC Bournemouth) and Mesut Özil (Arsenal) emerge as top players across multiple models due to their well-rounded contributions beyond direct goal involvement. \vspace{-0.15cm}
    \item \textit{Improved Contextual Understanding:} TransGoalNet, leveraging its transformer-based architecture, effectively models long-range dependencies, ensuring that off-the-ball movements and indirect contributions are properly valued. \vspace{-0.15cm}
\end{itemize}
These results reinforce the validity of the proposed approaches in soccer analytics, demonstrating their ability to uncover pivotal players beyond traditional performance evaluation methods. 

\textbf{Future Directions.}
Building upon this research, several extensions could further enhance the framework such as real-time adaptation --e.g., incorporating live match data to allow dynamic, real-time player performance assessments-- multimodal data tntegration --e.g., combining event-based match data with tracking data, physiological metrics, and tactical formations to refine player valuation-- and cross-league generalization.


\bibliography{example_paper}
\bibliographystyle{icml2025}

\newpage
\appendix
\onecolumn

\section{Ablation Studies}

This subsection examines the impact of temporal features (\( k \)-values) on model performance, focusing on comparisons across Baseline, GATGoalNet, and TransGoalNet models. Metrics such as Mean Absolute Error (MAE), Mean Squared Error (MSE), and their combined losses (train loss and validation loss) are analyzed for both training and validation datasets. 
The results also identify a ``favorable'' \( k \)-value for each model based on empirical loss.

\textbf{Impact of Temporal Features (\( k \)-Values).}
Tables~\ref{tab:ablation_k_values_mae} and ~\ref{tab:ablation_k_values_mse_loss} summarize the performance metrics for different \( k \)-values across the three models. The results indicate that while increasing the temporal context (higher \( k \)-values) generally improves performance, most models achieve their best results at \( k=7 \), with performance deteriorating at \( k=9 \).

\begin{table*}[!htp]
    \centering
    \begin{tabular}{l|ccccc|ccccc}
        \hline
        \textbf{Model} & \multicolumn{5}{c|}{\textbf{Train MAE}} & \multicolumn{5}{c}{\textbf{Val MAE}} \\
        \textbf{\( k \)} & 1 & 3 & 5 & 7 & 9 & 1 & 3 & 5 & 7 & 9 \\
        \hline
        Baseline      & 0.0198 & 0.0175 & 0.0290 & 0.0263 & 0.0352 & 0.0103 & 0.0082 & 0.0227 & 0.0139 & 0.0187 \\
        GATGoalNet    & 0.0106 & 0.0103 & 0.0123 & 0.0084 & 0.0152 & 0.0075 & 0.0082 & 0.0133 & 0.0066 & 0.0095 \\
        TransGoalNet  & 0.0043 & 0.0045 & 0.0033 & 0.0048 & 0.0065 & 0.0030 & 0.0031 & 0.0030 & 0.0031 & 0.0042 \\
        \hline
    \end{tabular}
    \caption{Training and validation MAE for varying \( k \)-values across Baseline, GATGoalNet, and TransGoalNet models.}
    \label{tab:ablation_k_values_mae}
\end{table*}

\begin{table*}[!htp]
    \centering
    \begin{tabular}{l|ccccc|ccccc}
        \hline
        \textbf{Model} & \multicolumn{5}{c|}{\textbf{Train MSE}} & \multicolumn{5}{c}{\textbf{Val MSE}} \\
        \textbf{\( k \)} & 1 & 3 & 5 & 7 & 9 & 1 & 3 & 5 & 7 & 9 \\
        \hline
        Baseline      & 0.0304 & 0.0221 & 0.0278 & 0.0862 & 0.1400 & 0.0104 & 0.0058 & 0.0074 & 0.0197 & 0.0380 \\
        GATGoalNet    & 0.0016 & 0.0005 & 0.0035 & 0.0022 & 0.0202 & 0.0017 & 0.0006 & 0.0045 & 0.0020 & 0.0110 \\
        TransGoalNet  & 0.0003 & 0.0003 & 0.0001 & 0.0004 & 0.0063 & 0.0001 & 0.0002 & 0.0001 & 0.0002 & 0.0043 \\
        \hline
    \end{tabular}
    \caption{Training and validation MSE for varying \( k \)-values across Baseline, GATGoalNet, and TransGoalNet models.}
    \label{tab:ablation_k_values_mse}
\end{table*}

\begin{table*}[!htp]
    \centering
    \begin{tabular}{l|ccccc|ccccc}
        \hline
        \textbf{Model} & \multicolumn{5}{c|}{\textbf{Train Loss (MAE + MSE)}} & \multicolumn{5}{c}{\textbf{Val Loss (MAE + MSE)}} \\
        \textbf{\( k \)} & 1 & 3 & 5 & 7 & 9 & 1 & 3 & 5 & 7 & 9 \\
        \hline
        Baseline      & 0.0502 & 0.0396 & 0.0568 & 0.1125 & 0.1752 & 0.0207 & 0.0140 & 0.0300 & 0.0336 & 0.0567 \\
        GATGoalNet    & 0.0122 & 0.0108 & 0.0158 & 0.0106 & 0.0354 & 0.0092 & 0.0088 & 0.0178 & 0.0086 & 0.0205 \\
        TransGoalNet  & 0.0046 & 0.0048 & 0.0034 & 0.0052 & 0.0128 & 0.0031 & 0.0033 & 0.0031 & 0.0033 & 0.0085 \\
        \hline
    \end{tabular}
    \caption{Training and validation combined loss (MAE + MSE) for varying \( k \)-values across Baseline, GATGoalNet, and TransGoalNet models.}
    \label{tab:ablation_k_values_mse_loss}
\end{table*}

\textbf{Insights on Temporal Context.}
The results demonstrate that while higher \( k \)-values provide additional temporal context, they can also introduce overfitting or increased noise, as seen in the deterioration of performance at \( k=9 \). Most models achieve their best performance at \( k=7 \), except TransGoalNet, which performs best at \( k=5 \) when evaluated based on loss.





Overall, these findings emphasize the importance of carefully selecting \( k \)-values to optimize model performance while considering computational efficiency and potential overfitting.


\section{GoalNet with Graph Attention Networks (GATNet)}

This section presents a variation of the GoalNet framework that replaces standard graph convolutional layers with Graph Attention Network (GAT) layers \citep{velickovic2018gat}. 
This modification introduces a data-driven mechanism for weighting player interactions, allowing the model to focus on the most influential relations in the team’s passing network and event dynamics. 
By doing so, we aim to refine the identification and credit assignment of pivotal players, especially in scenarios where certain passes, tackles, or sequences bear disproportionate importance to the resulting expected threat (xT) changes.

In contrast, in a vanilla GCN, the embedding update for a node is computed by summing or averaging the features of its neighbors with fixed weights derived from the graph’s structure. 
In GAT, instead of using the adjacency matrix directly, the model employs attention mechanisms to learn dynamic and context-dependent weights for each edge. 
More concretely, each node pair’s features are fed into a learned function that computes attention coefficients indicating how strongly these nodes should influence each other.

\subsection{Graph Representation of Events}

Similar to the original GoalNet, every soccer event is modeled as a graph \( G=(V,E) \), with players as nodes \( v \in V \) and their interactions as edges \( e \in E \). Each node encapsulates the same set of attributes described previously—player-specific statistics, positional data, and historical performance metrics. Edges capture player-to-player interactions (e.g., passes, tackles) and encode information such as event type, outcome, spatial coordinates of the action, and the corresponding changes in xT values.

In this GAT-based extension (GATNet), the node and edge feature sets remain identical to the original formulation to ensure consistency and comparability. The critical difference lies in how these features are integrated during message-passing steps.

\subsection{Node and Edge Attributes}

\textbf{Node attributes.} Each node \( v \) still represents a player, characterized by features such as pass accuracy, interceptions, and key passes. We denote the node feature vector as \(\mathbf{x}_v\). These features remain unchanged from the original GoalNet approach, ensuring that the player descriptors capture performance comprehensively.

\textbf{Edge attributes.} As before, edges represent interactions including passes, tackles, and transitions. The edge feature vector \(\mathbf{e}_{uv}\) between players \( u \) and \( v \) encodes attributes like pass start/end coordinates, event result, and xT change. While GAT primarily focuses on node feature transformations, these edge attributes can still be incorporated as additional inputs or as part of a preprocessing step to enrich the node embeddings before the attention computation.

\subsection{GAT-Based Architecture}

In contrast to the standard GNN layers used by GoalNet, we now apply GAT layers to process the event graphs. Instead of computing node updates by a uniform averaging or summation of neighbor features, GAT layers assign learned, adaptive attention weights to each neighbor. This mechanism allows the model to highlight crucial edges (e.g., decisive forward passes) over less impactful interactions (e.g., a routine backward pass in a non-threatening area).

\textbf{Attention mechanism.} For each node \( v \) and its neighbor \( u \in N(v) \), we compute an attention coefficient \(\alpha_{uv}\) that captures the importance of node \( u \)’s features to node \( v \). Specifically, if \(\mathbf{h}^{(l)}_v\) is the hidden embedding of node \( v \) at layer \( l \), and \( \mathbf{W} \) is a linear projection matrix, then we first transform node features and then apply a shared attention vector \(\mathbf{a}\):
\[
\alpha_{uv} = \frac{\exp\left(\text{LeakyReLU}\left( \mathbf{a}^\top [\mathbf{W} \mathbf{h}_u^{(l)} \| W\mathbf{W}\mathbf{h}_v^{(l)}]\right)\right)}{\sum_{w \in N(v)} \exp\left(\text{LeakyReLU}\left(\mathbf{a}^\top [\mathbf{W} \mathbf{h}_w^{(l)} \| \mathbf{W} \mathbf{h}_v^{(l)}]\right)\right)},
\]
where \(\|\) denotes concatenation. These attention coefficients \(\alpha_{uv}\) determine how much of \( u \)'s information node \( v \) will absorb. Nodes linked by critical interactions—those that significantly alter the match’s threat dynamics—will naturally receive higher attention scores once the model learns their importance.

\textbf{Multi-head attention.} To enhance the robustness of the learned attention and capture diverse aspects of player roles and interactions, we employ multi-head attention. Multiple sets of \((\mathbf{W},\mathbf{a})\) parameters run in parallel, each producing a separate embedding. These are then concatenated or averaged to yield the final updated node embedding:
\[
\mathbf{h}_v^{(l+1)} = \sigma\left(\bigg\Vert_{m=1}^M \sum_{u \in N(v)} \alpha_{uv}^{(m)} \mathbf{W}^{(m)} \mathbf{h}_u^{(l)}\right),
\]
where \(M\) is the number of attention heads, and \(\alpha_{uv}^{(m)}\) are the attention coefficients for head \(m\).

\subsection{Global Pooling and Output Layers}

After applying one or more GAT layers, we perform global mean pooling over the node embeddings \(\mathbf{h}_v\) to obtain a single graph-level embedding \(\mathbf{z}\). This aggregated vector represents the event as a whole, incorporating not just who was involved but also how important their interactions were, according to the learned attention weights.

A set of fully connected layers then predict \(\Delta xT\) from \(\mathbf{z}\). The training objective remains the same, using a Mean Squared Error (MSE) or Mean Absolute Error (MAE) loss to align predicted \(\Delta xT\) with the actual values derived from the dataset.

\subsection{xT Attribution to Players}

Similar to the original GoalNet approach, once we have a trained GAT-based model, we use the learned node embeddings to distribute the \(\Delta xT\) among the involved players. However, the GAT framework provides more structured embeddings. Because the attention mechanisms highlight the edges and neighbors that most strongly influence a node’s representation, the resulting node embeddings may more clearly reflect critical player contributions.

We extract node embeddings from the last GAT layer and compute their magnitudes to gauge the relative importance of each player during the event. By normalizing these magnitudes and multiplying by the predicted \(\Delta xT\), we allocate the xT credit proportionally, ensuring that players responsible for high-impact interactions get properly recognized.

\subsection{Temporal Context and Additional Considerations}

As with the original GoalNet, we can incorporate a temporal window of previous \( k \) events to inform the model about recent plays. GAT layers, when applied to each event’s graph, enable a richer encoding of the current tactical snapshot. Temporal modeling via a sliding window remains identical, but each snapshot is now processed with attention-based weighting of edges. This may help the model focus on recent key interactions—like a crucial pass interception just before a scoring opportunity—over older, less relevant events.

Additionally, integrating GAT layers does not preclude the use of other enhancements. For example, GAT layers could be combined with multi-task learning or hierarchical representations. Yet even as a standalone improvement, the move from static neighbor aggregation to adaptive attention weighting promises a finer-grained identification of the pivotal, if sometimes understated, contributions of players who facilitate and orchestrate team play.

In summary, GATNet preserves the original GoalNet pipeline—graph construction, node and edge feature extraction, \(\Delta xT\) prediction, and player-level xT distribution—while substituting standard graph convolutions with attention-driven message passing. This refinement highlights key edges and nodes, potentially leading to more accurate and interpretable measurements of player impact on match outcomes.

\section{GoalNet with Graph Transformers (TransGoalNet)}

This section presents an alternative extension to the baseline GoalNet framework using Graph Transformers. While GoalNet models interactions using local aggregations of neighbor information, Graph Transformers \citep{dwivedi2020generalization} aim to capture long-range dependencies and global structural patterns in a graph. In other words, unlike GATs, which focus on assigning attention weights to a node’s immediate neighbors, Graph Transformers employ global self-attention across the entire set of nodes. This holistic view may better reflect the multi-faceted and evolving nature of soccer, where off-ball positioning and distant player roles influence the expected threat (xT) dynamics.

\subsection{Graph Representation of Events}

As before, each event in a soccer match is represented as a graph \( G = (V,E) \), where nodes represent players and edges represent their interactions. The same node and edge attributes described in the baseline GoalNet model are retained. Nodes carry player-specific performance indicators and role-based statistics, while edges encode event-related features, including the spatial coordinates of passes or tackles and the associated xT change. The underlying graph structure, thus, remains consistent: a snapshot of how players coordinate and influence scoring probabilities at a specific event in time.

In TransGoalNet, the novelty lies not in the graph construction but in how node features are transformed and aggregated. Graph Transformers consider all node pairs at once, enabling the model to capture patterns extending beyond immediate connections.

\subsection{Node and Edge Attributes}

\textbf{Node attributes.} Each player node \( v \) includes the same set of attributes as in GoalNet: performance metrics, positional info, passing accuracies, and defensive/offensive contributions. The initial node feature vector \(\mathbf{x}_v\) remains identical, ensuring compatibility and fair comparisons with other variants.

\textbf{Edge attributes.} Similarly, edges store event details: pass distances, event success/failure, xT values, and event types. However, unlike GATs or GCNs, Graph Transformers do not explicitly rely on adjacency for neighbor-based filtering. The edges serve more as features that can inform positional or relational encodings added to nodes, potentially affecting the attention mechanism.

\subsection{Graph Transformer-Based Architecture}

At the core of Graph Transformers is a self-attention mechanism applied not just over a node’s neighbors, but over all nodes in the graph. Each layer performs a global attention operation, allowing any node to directly attend to all other nodes. This ensures that even distant players—perhaps a fullback positioned far across the pitch—can influence the representation of a forward if that positional setup holds strategic importance.

\textbf{Positional and structural encodings.} Unlike linear sequences in language models, graphs do not have a natural order. For soccer analytics, we can inject information about player positions, roles, or even centrality metrics as positional encodings. For example, one might encode a player’s on-field coordinates as a 2D positional embedding or incorporate learned embeddings representing the player’s role (e.g., defender, midfielder, forward). These embeddings are added to the initial node features to provide the transformer layers with spatial and tactical context.

\textbf{Global self-attention.} Each transformer layer computes query, key, and value matrices \( (\textbf{Q},\textbf{K},\textbf{V}) \) from the node embeddings \(\mathbf{H}^{(l)}\) at layer \(l\):
\[
\textbf{Q} = \mathbf{H}^{(l)} \textbf{W}_\textbf{Q},\quad \textbf{K} = \mathbf{H}^{(l)} \textbf{W}_\textbf{K},\quad \textbf{V} = \mathbf{H}^{(l)} \textbf{W}_\textbf{V}.
\]
Here, \( \textbf{W}_\textbf{Q}, \textbf{W}_\textbf{K}, \textbf{W}_\textbf{V} \) are learnable parameters. Attention coefficients are then computed for every pair of nodes:
\[
\text{Att}(\textbf{Q},\textbf{K},\textbf{V}) = \text{softmax}\left(\frac{\textbf{QK}^\top}{\sqrt{d}}\right)\textbf{V},
\]
where \( d \) is a scaling factor (the dimension of the embeddings). Unlike GAT, which focuses on a node’s immediate neighbors, Graph Transformers yield a dense attention matrix, enabling interactions between all node pairs. Multi-head attention is again employed to provide multiple “views” of the global player configuration, focusing on different strategic aspects.

\textbf{Layer normalization and feed-forward layers.} Each transformer layer also includes layer normalization and position-wise feed-forward networks:
\[
\mathbf{H}^{(l+1)} = \text{LayerNorm}\left(\mathbf{H}^{(l)} + \text{Att}(\textbf{Q},\textbf{K},\textbf{V})\right),
\]
\[
\mathbf{H}^{(l+1)} = \text{LayerNorm}\left(\mathbf{H}^{(l+1)} + \text{FFN}(\mathbf{H}^{(l+1)})\right),
\]
where \(\text{FFN}\) is a feed-forward network applied to each node embedding independently. This ensures stable training and non-linear transformations of the node features.

\subsection{Global Pooling and Output Layers}

As in GoalNet and GAT-based solutions, once the transformer layers produce updated node embeddings, we apply a global pooling operation (e.g., mean pooling):
\[
\mathbf{z} = \frac{1}{|V|}\sum_{v \in V}\mathbf{h}_v^{(L)},
\]
where \(\mathbf{h}_v^{(L)}\) is the node embedding of player \( v \) after \(L\) transformer layers. The pooled embedding \(\mathbf{z}\) captures the event-level representation in a highly context-aware manner, incorporating both local and global structural cues.

Fully connected layers map \(\mathbf{z}\) to the predicted \(\Delta xT\), and the training objective remains minimizing a loss (e.g., MSE) between predicted and actual \(\Delta xT\).

\subsection{xT Attribution to Players}

Once trained, we use the final node embeddings from the transformer layers to allocate \(\Delta xT\) across players. Because the transformer layers consider all player relationships at once, each node embedding should encapsulate both local involvements and the global context in which a player operates. Normalizing the embedding magnitudes for players involved in the event and multiplying by \(\Delta xT\) distributes credit in a manner that accounts not only for direct interactions but also strategic placements and roles.

This may highlight subtle but crucial contributions: a defensive midfielder’s positioning may indirectly enable a forward’s xT increase by forcing opponents into disadvantageous positions, something less easily captured if only local neighborhoods are considered.

\subsection{Temporal Context and Additional Considerations}

Temporal context integration (via a sliding window of past events) remains the same as in GoalNet. For each event snapshot, a Graph Transformer processes the graph independently. By maintaining multiple historical embeddings, the model can still incorporate temporal dependencies, potentially via a separate temporal modeling component.

Furthermore, the Graph Transformer’s ability to handle global context could improve performance over GATs or the simple GoalNet in scenarios where complex, long-range patterns matter. For example, if a right-back’s run on the flank shifts the opposing team’s formation, indirectly increasing \(\Delta xT\) for a forward positioned on the opposite side, a Graph Transformer may detect this relationship more easily than a GAT, which relies on neighbor-level attention. Similarly, the baseline GoalNet (using simple GCN layers) primarily focuses on local structures and may miss such global tactical patterns.

In theory, Graph Transformers can outperform GATs or simpler GNNs when the game’s complexity arises from strategic formations and long-distance player influences. By not restricting message passing to local neighborhoods, Graph Transformers can better model the spatial and positional interplay among all players on the field. As a result, this comprehensive global attention can reveal indirect and nuanced factors affecting \(\Delta xT\), potentially leading to more accurate predictions and more insightful player evaluations.

In summary, TransGoalNet maintains the original pipeline for building the event graph and predicting \(\Delta xT\), but infuses a global view of the game’s geometry and relational patterns via Graph Transformers. This approach’s advantage lies in its capacity to handle complex, globally interacting structures, providing richer insights into the hidden orchestrators of team performance.

\end{document}